\documentclass[10pt,twocolumn,letterpaper]{article}
\usepackage[pagenumbers ]{cvpr} 

\definecolor{cvprblue}{rgb}{0.21,0.49,0.74}
\usepackage{color}
\usepackage{xcolor}
\usepackage{graphicx}
\usepackage{booktabs}
\usepackage{amsmath, amsthm, amsfonts, amssymb}
\usepackage{mathrsfs}
\usepackage{textcomp}
\usepackage{epstopdf}
\usepackage{multirow}
\usepackage{wrapfig}
\usepackage{booktabs} 
\usepackage[ruled]{algorithm2e} 
\DontPrintSemicolon
\SetKwComment{Comment}{// }{}
\usepackage{algorithmicx}  
\usepackage{algpseudocode}
\usepackage{ragged2e}
\usepackage[normalem]{ulem}
\usepackage{bm}
\usepackage{makecell}
\usepackage[pagebackref,breaklinks,colorlinks,allcolors=cvprblue]{hyperref}
\usepackage{cleveref}
\PassOptionsToPackage{capitalize}{cleveref}

\SetAlFnt{\small}
\SetAlCapFnt{\small}
\SetAlCapNameFnt{\small}
\SetAlCapHSkip{0pt}

\definecolor{green}{rgb}{0, 0.5, 0}
\definecolor{orange}{rgb}{0.6, 0.3, 0.1}
\definecolor{red}{rgb}{1.0, 0.0, 0.0}
\definecolor{teal}{rgb}{0.0, 0.4, 0.4}
\definecolor{purple}{rgb}{0.65,0,0.65}
\definecolor{saffron}{rgb}{0.95,0.75,0.2}
\definecolor{turquoise}{rgb}{0.0,0.5,0.5}
\definecolor{brown}{rgb}{0.5, 0.16, 0.16}
\definecolor{brickred}{rgb}{.6, .2, .1}
\definecolor{coral}{rgb}{1,0.45,0.33}
\definecolor{newcolor}{rgb}{.8,.349,.1}
\definecolor{ceruleanblue}{rgb}{0.16, 0.32, 0.75}

\definecolor{algcomment}{RGB}{75,115,116}

\newcommand{\zs}[1]{{\color{black} #1}}

\newcommand{\dl}[1]{{\color{black} #1}}
\newcommand{\gz}[1]{{\color{black} #1}}

\title{Cycle-Consistent Tuning for Layered Image Decomposition}

\author{
Zheng Gu$^{1}$\quad
Min Lu$^{1}$\quad
Zhida Sun$^{1}$\quad
Dani Lischinski$^{2}$\quad
Daniel Cohen-Or$^{3}$\quad
Hui Huang$^{1}$\thanks{Corresponding author.} \\
$^{1}$Shenzhen University \quad
$^{2}$Hebrew University of Jerusalem \quad
$^{3}$Tel Aviv University \\
{\tt\small \{guzheng.szu, lumin.vis, zhdsun, danix3d, cohenor, hhzhiyan\}@gmail.com}
}

\begin{document}

\twocolumn[{
\maketitle
\begin{center}
    \includegraphics[width=\textwidth]{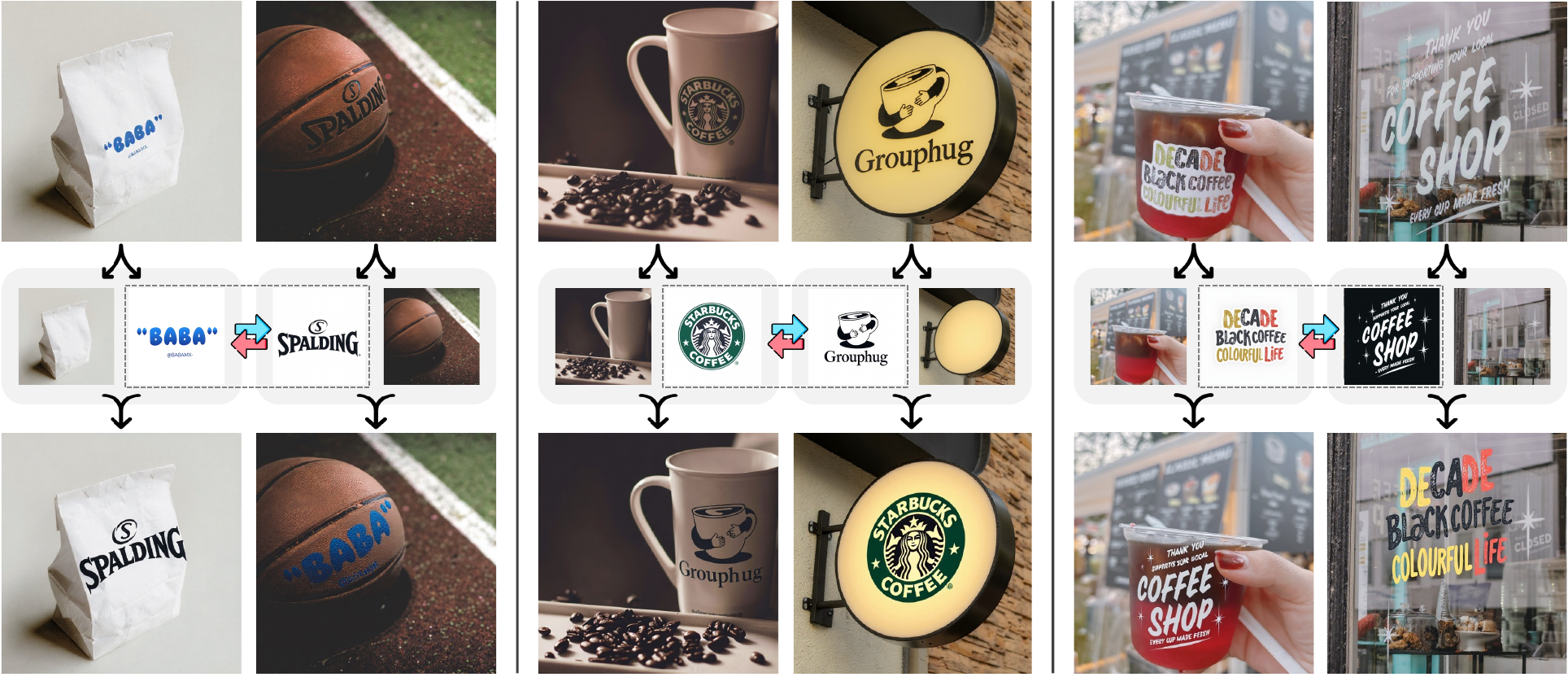}
    \captionof{figure}{Our method learns to disentangle overlaid logos from their supporting surfaces and recompose them seamlessly onto other objects.
Each example shows two input photographs of objects with distinct logos. We first decompose each image into its logo and object layers, and then cross-compose the separated logos onto the other objects. These results demonstrate accurate separation and faithful re-integration across challenging non-linear cases involving complex geometry, lighting, and viewpoint changes.}
    \label{fig:teaser}
\end{center}
}]

\begingroup
\renewcommand{\thefootnote}{\fnsymbol{footnote}}
\footnotetext[1]{Corresponding author}
\endgroup

\begin{abstract}

Disentangling visual layers in real-world images is a persistent challenge in vision and graphics, as such layers often involve non-linear and globally coupled interactions, including shading, reflection, and perspective distortion. In this work, we present an in-context image decomposition framework that leverages large diffusion foundation models for layered separation. We focus on the challenging case of logo-object decomposition, where the goal is to disentangle a logo from the surface on which it appears while faithfully preserving both layers. Our method fine-tunes a pretrained diffusion model via lightweight LoRA adaptation and introduces a cycle-consistent tuning strategy that jointly trains decomposition and composition models, enforcing reconstruction consistency between decomposed and recomposed images. This bidirectional supervision substantially enhances robustness in cases where the layers exhibit complex interactions. Furthermore, we introduce a progressive self-improving process, which iteratively augments the training set with high-quality model-generated examples to refine performance. Extensive experiments demonstrate that our approach achieves accurate and coherent decompositions and also generalizes effectively across other decomposition types, suggesting its potential as a unified framework for layered image decomposition. Project website is \href{https://vcc.tech/research/2026/ImgDecom}{here}.

\end{abstract}
\section{Introduction}

Image decomposition has long been a challenging problem in computer vision and computer graphics, aiming to factorize images into semantically or physically meaningful layers or components. Classic approaches such as intrinsic decomposition~\cite{careaga2023intrinsic,careaga2024colorful,zeng2024rgb} separate reflectance from shading, typically relying on \gz{explicit} priors and rigid task formulations. More recent methods leverage alpha-channel representations~\cite{zhang2024transparent} for layered image decomposition~\cite{yang2025generative,wang2025diffdecompose,suzuki2025layerd}, yielding structured outputs through additive layer compositing.

These techniques, however, are largely confined to settings where components interact linearly (e.g., via alpha blending~\cite{zhang2024transparent,suzuki2025layerd}). In contrast, isolating a logo from a product photographed under non-frontal viewpoints involves highly non-linear and globally coupled interactions driven by shading, perspective distortion, surface reflectance, and material-dependent appearance~\cite{li2025assetdropper}. Such cases cannot be resolved by local or patch-based analysis alone; they require non-local reasoning and semantic understanding of what constitutes an object versus an overlaid element. Addressing this challenge calls for data-driven priors that capture scene- and object-level context, as encoded by modern foundation models~\cite{rombach2022high,flux2024}.

\dl{
In this work, we leverage the representational power of large foundation models to tackle logo-object image decomposition. Rather than relying on handcrafted priors or local statistics~\cite{careaga2024colorful}, we adopt a pretrained \gz{image inpainting} diffusion model to separate an overlaid logo from its supporting object, producing (i) a rectified logo layer (fronto-parallel and largely illumination-invariant) and (ii) a ``clean'' object image. We realize this adaptation via a lightweight LoRA fine-tuning~\cite{hu2022lora} tailored to this task. \gz{Our training follows the In-Context Learning (ICL) paradigm~\cite{huang2024context}, where the supervision is presented in a single \gz{three-panel} grid image.} This encourages the model to internalize the operation of removing or isolating overlaid elements while preserving the underlying structure. Importantly, our method is not a training-free, one-shot scheme~\cite{bar2022visual,gu2024analogist} conditioned on example pairs during inference; instead, we adopt the data-driven route and train on a carefully curated dataset to ensure the outputs remain contextual, consistent, and faithful to the input. This design imparts task-specific decomposition capabilities while retaining the broad generalization of the pretrained model, with performance that scales with the availability and quality of training data.


To make this training effective and \gz{robust}, we introduce a coupled decomposition–composition framework with cycle consistency. The decomposition module predicts two outputs from a composite input: the rectified logo and the logo-free object. A complementary composition module then reconstructs the original image from these predicted components. A cycle-consistency loss enforces agreement between the reconstruction and the input, allowing the two modules to supervise each other and reducing the need for densely annotated ground truth. Practically, we organize supervision as triplets (composite, logo, clean object) when available, and fall back on the cycle signal when ground-truth layers are incomplete or imperfect. This joint training substantially stabilizes learning and improves fidelity in the presence of real-world non-linearities.


We further propose a self-improving data loop that progressively enlarges and refines the training set. \gz{Starting from a small seed of annotated triplets, we train an initial LoRA, use it to generate additional candidate labeled triplets, and select high-quality results to refine the LoRA and yield more reliable labeled data. After that, we train the cycle-consistent model and use it to propose additional decompositions on unlabeled images, selecting high-quality results via automated filters and simple human-in-the-loop checks.} The accepted results are then added back into the training pool for subsequent rounds. Combined with our cycle-consistent objective, this bootstrapping strategy steadily improves robustness and semantic accuracy across diverse viewing conditions, materials, and lighting.

We validate the effectiveness of the approach through extensive experiments, as illustrated in \Cref{fig:teaser}. While our primary focus is logo–object decomposition, we also demonstrate the generality of the framework by applying it to two distinct problems: foreground–background separation and ambient-level decomposition of albedo and lighting. These results suggest a general paradigm for image decomposition that can handle complex, non-linear, and semantically coupled interactions well beyond the specific application studied in this work.
}

\section{Related Work}
\label{sec:related}

\subsection{Diffusion Models}

Diffusion models \cite{ho2020denoising} have emerged as the leading framework for high-fidelity visual generation in recent years. They model complex data distributions through a progressive denoising process \cite{song2019generative}. Since the Stable Diffusion \cite{rombach2022high} and the Diffusion Transformer \cite{peebles2023scalable}, these approaches have advanced text-to-image synthesis \cite{betker2023improving}, image translation \cite{li2023bbdm}, and controllable image editing \cite{meng2021sdedit,zhang2023adding}.
Recent variants such as the FLUX family~\cite{flux2024,labs2025flux} further improves contextual awareness, enabling localized and spatially consistent manipulations.
Building on this progress, we extend diffusion models to image decomposition, where a single input is split into distinct components.


\begin{figure*}[!t] 
\centering \includegraphics[width=\textwidth]{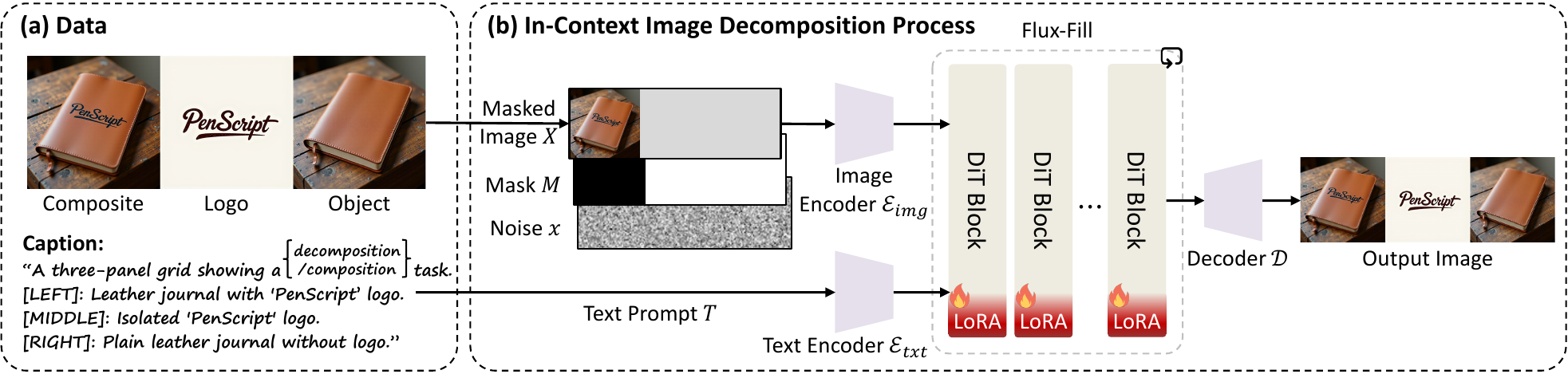} \caption{Overview of the image decompostion framework. Given a composite image, the model receives a masked input, a binary mask indicating the logo region, and a noise latent, and predicts both the isolated logo and the clean object. The process is implemented by tuning a LoRA on top of Flux-Fill. The composition scheme is similar but uses a complementary mask to produce the composite image.} 
\label{fig:decomp_process} 
\end{figure*}

\subsection{Visual In-Context Learning}


In-Context Learning (ICL)~\cite{dong2024survey}, originated from large language models (LLMs), indicates the ability of pretrained models to adjust their behavior by leveraging contextual signals~\cite{kojima2022large}. Early works of Visual ICL focus on discriminative tasks such as image segmentation~\cite{wang2023seggpt,wang2023images,zhang2023makes}. Some later works explore generative visual ICL, such as conditional image generation~\cite{bar2022visual,xu2023improv,wang2023context} and example-based image editing~\cite{gu2024analogist,li2025visualcloze,chen2025transfer}. More recently, researchers have found that existing Diffusion Models possess inherent in-context abilities during their generation process~\cite{huang2024context}, which enables applications such as image editing through text instructions~\cite{labs2025flux,zhang2025context}. Although these techniques demonstrate strong capability, they are mostly confined to \textit{single-input-single-output} mappings. 
In this work, we unlock the \textit{single-input-multi-output} potential of visual ICL. 
We achieve this via a simple but effective cycle-consistent tuning strategy, which decomposes and recomposes with contextual information to enforce mutual correspondence between layers.

\subsection{Asset Extraction via Generative Models}

Recent work has explored the use of generative models for extracting reusable assets directly from images. 
Affara et al.~\cite{affara2016large} explored asset extraction for urban street assets through rectified object proposals.
RRM~\cite{gomez2024rrm} introduces radiance-guided material extraction to obtain relightable 3D assets.
AssetDropper~\cite{li2025assetdropper} formulates this problem through conditional image-to-image translation, learning to isolate standardized assets from user-specified regions. It employs a reward-driven optimization process that aligns the extracted assets with visual priors.
In contrast, we view this problem from a more general layered decomposition angle, where the decomposition is learned jointly with another composition process. Rather than relying on explicit masks or fixed extraction views, our approach infers the layered structure directly from context, enabling decomposition that preserves layers faithfully beyond asset extraction.


\section{Method}
\label{sec:method}



In this work, we present a framework for image decomposition using large generative models. Leveraging the contextual understanding capabilities of Diffusion Transformers, we fine-tune a pretrained image inpainting model with LoRA to specialize in separating components from single composite images (\Cref{sec:preliminary}). To ensure consistency and plausibility between input and decomposed layers, we jointly train a complementary composition model under a cycle-consistency constraint (\Cref{sec:cycle_consistency}). Furthermore, \gz{we adopt a progressive data collection with a self-improvement strategy} that iteratively expands the training using generated pseudo-decompositions filtering (\Cref{sec:self_improv}). 





\subsection{Preliminaries}
\label{sec:preliminary}

\paragraph{Image Inpainting with DiTs.}
Our method is built upon FLUX.1-Fill-dev~\cite{flux2024}, a Diffusion Transformer model designed for image inpainting. Given a masked image $X$, a binary mask $M$, and noise latent $x$, the model iteratively refines the masked region through a conditional Transformer guided by a text prompt $T$. As shown in ~\Cref{fig:decomp_process}, at each timestep $t$, the forward process can be formulated as:
\begin{equation}
    x_{t-1} = \phi\left(v_{\theta}\left(\left[x_t,\mathcal{E}_{img}\left(X\right),M\right], t, \mathcal{E}_{txt}\left(T\right)\right)\right),
\end{equation}
where $v_{\theta}$ denotes the model, $\mathcal{E}_{img}$ and $\mathcal{E}_{txt}$ are image and text encoders, and $\phi$ is the flow-matching~\cite{esser2024scaling} scheduler. After denoising, the latent $x_0$ is fed into the VAE decoder $\mathcal{D}$ to reconstruct the complete image.

\paragraph{Low-Rank Adaptation (LoRA).}
We adopt Low-Rank Adaptation (LoRA)~\cite{hu2022lora} to efficiently fine-tune the Diffusion Transformer. Given a parameter matrix $W \in \mathbb{R}^{d \times k}$, LoRA learns the following matrices:
\begin{equation}
    W' = W + UV,
\end{equation}
where $U \in \mathbb{R}^{d \times r}$ and $V \in \mathbb{R}^{r \times k}$ are LoRA parameters ($r \ll d$). This enables lightweight adaptation while preserving the pretrained model's capacity.

Specifically, to finetune the inpainting model for decomposition, flow matching loss is utilized to encourage the Transformer to predict the velocity field of the forward process as follows:
\begin{equation}
    \mathcal{L}_{rec}=\mathbb{E}_{x,t}\left[ {\left\| v_{\theta} \left( x_t, M, t, \tau \right) - \frac{\partial x_t}{\partial x} \right\|}^2_2 \right],
\end{equation}
where ${\theta}=(U,V)$. $M$ is set to zeros in the regions to be preserved and ones in the regions to be inpainted.

\begin{figure}[!t] 
\centering \includegraphics[width=\linewidth]{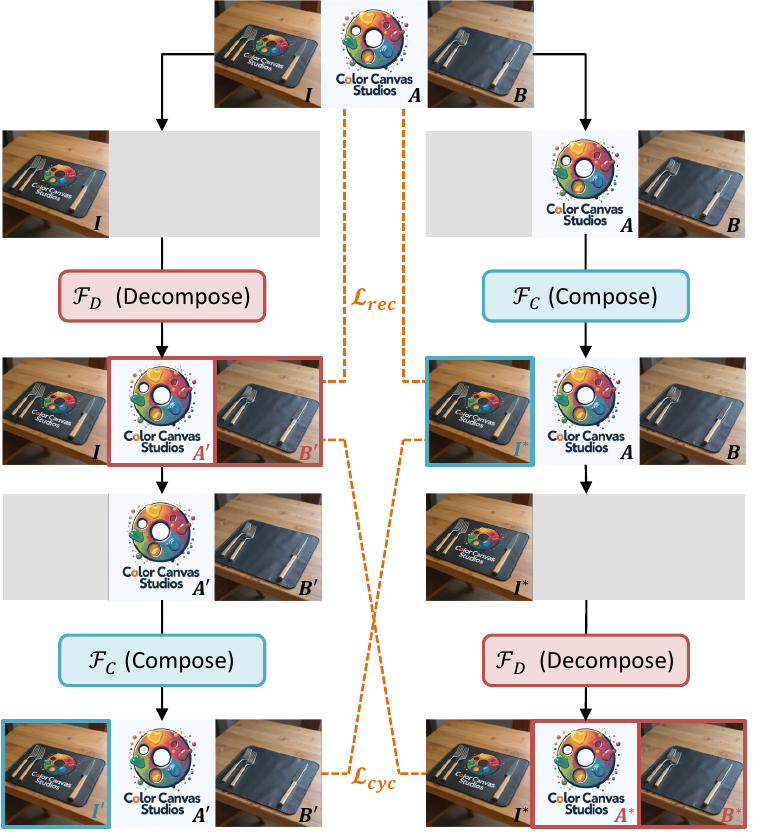} \caption{Illustration of our cycle-consistent training. The model jointly learns decomposition and composition, ensuring the decomposed layers recompose into the original image and vice versa.
}
\label{fig:cycle_consistency} \end{figure}

\subsection{Cycle Consistent Decomposition}
\label{sec:cycle_consistency}


Typically, image decomposition is an ill-posed problem since the number of unknowns exceeds the number of inputs. Although we can collect ground truth decomposition images to some extent, the supervision is still insufficient to constrain the model.
To address this challenge, we introduce a cycle consistency constraint to provide additional regularization to the solution space. Unlike decomposition, composition is a well-defined and deterministic process. Therefore, by jointly learning decomposition and composition in a cyclic manner, we enable the model to leverage complementary knowledge from both tasks, thereby stabilizing the training and reducing ambiguity.


As shown in ~\Cref{fig:cycle_consistency}, the model is given an example triplet of images $\left\langle I, A, B \right\rangle$, where $I$ denotes the combination of $A$ and $B$. Note that this combination can be either linear alpha blending or more complex nonlinear interactions, such as logo-object composition. Our goal is to learn the following two functions at the same time:
\begin{equation}
    \left\{
    \begin{aligned}
    &\mathcal{F}_{D}(I) = \left\langle A, B \right\rangle, \\
    &\mathcal{F}_{C}\left( \left\langle A, B \right\rangle \right) = I,
    \end{aligned}
    \right.
\end{equation}
where $\mathcal{F}_{D}$ is the decomposition function and $\mathcal{F}_{C}$ is the composition function.
To learn both functions, during each training step, we operate decomposition and composition symmetrically to supervise each other in two parallel tracks: 

\begin{enumerate}
    \item Starting from $I$, first predict $\left\langle A', B' \right\rangle = \mathcal{F}_{D}(I)$, then recompose them into $I'=\mathcal{F}_{C}(\left\langle A', B' \right\rangle)$;
    \item Starting from $A$ and $B$, first predict $I^*=\mathcal{F}_{C}(\left\langle A, B \right\rangle)$, then decompose it into $\left\langle A^*, B^* \right\rangle = \mathcal{F}_{D}(I^*)$.
\end{enumerate}

We employ a cycle consistency loss $\mathcal{L}_{cyc}$ to drive the above learning process:
\begin{equation}
\begin{split}
\mathcal{L}_{cyc} = \mathbb{E}_{x,t_1}\!\Big[
\|v_{\theta}(x_{t_1}^I,M_D,t_1,\tau_D)
- v_{\theta}(x_{t_1}^{I^*},M_D,t_1,\tau_D)\|_2^2
\Big] \\
+ \mathbb{E}_{x,t_2}\!\Big[
\|v_{\theta}(x_{t_2}^{\langle A,B\rangle},M_C,t_2,\tau_C)
- v_{\theta}(x_{t_2}^{\langle A',B'\rangle},M_C,t_2,\tau_C)\|_2^2
\Big],
\end{split}
\end{equation}
where $t_1$ and $t_2$ are two different timesteps. $\{M_D,\tau_D\}$ and $\{M_C, \tau_C\}$ are the binary mask and text tokens for decomposition and composition, respectively. In practice, the two functions share the same LoRA parameter space $\theta$, which enhances parameter efficiency and stabilizes the training.

\begin{figure*}[!t] 
\centering \includegraphics[width=\linewidth]{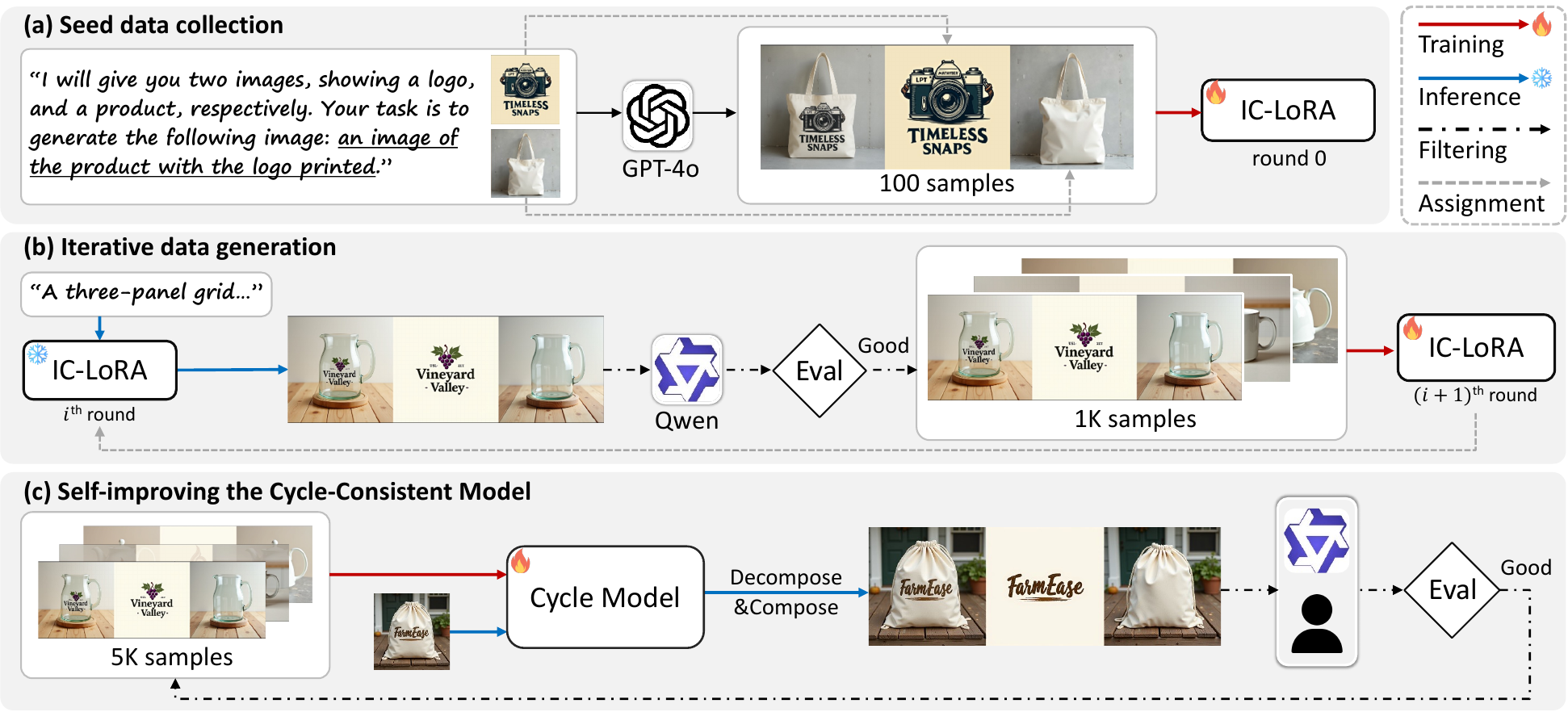} \caption{Illustration of progressive data collection. (a) We first collect a seed dataset to obtain an IC-LoRA as the initial data generator. (b) In each round, we select high-quality \zs{samples generated by the current LoRA and reintroduce them to the training set. (c) During the training of the cycle model, we use it to produce consistent data by decomposing an image, and then re-composing it}. High-quality recomposition samples are added back into the training set.} 
\label{fig:data_collection} \end{figure*}




\subsection{Progressive Data Collection}
\label{sec:self_improv}

Training high-quality decomposition models usually requires large paired datasets.
\zs{However, collecting such data for our logo–object decomposition is costly and hard to scale.}
\zs{To overcome this data scarcity, we propose a progressive self-improvement strategy that iteratively expands the training corpus using model-generated data.
}

\paragraph{Seed Data Collection.}
We initiate the process with a small seed dataset comprising 100 manually curated $\left\langle I, A, B \right\rangle$ triplets, with the assistance of GPT-4o~\cite{hurst2024gpt} (\Cref{fig:data_collection}(a)). This seed set is used to train an initial IC-LoRA~\cite{huang2024context}, which is then employed to generate a large candidate pool of new triplets from text prompts.

\zs{
\paragraph{Iterative Data Generation with IC-LoRA.}
Given the sparse initial data, the first-generation IC-LoRA exhibits instability and often fails to produce plausible decompositions. A naive expansion of the dataset with these low-quality samples would lead to error propagation. To address this, we generate data iteratively (\Cref{fig:data_collection}(b)). We use Qwen-VL~\cite{bai2025qwen2} to filter generated triplets based on visual plausibility and decomposition consistency. This filtered dataset is then used to train IC-LoRA for the next round. This iterative process of generation, filtering, and retraining progressively enhances the data generation stability.
}


\zs{
\paragraph{Self-improving the Cycle-Consistent Model.}
We apply a similar iterative self-improving methodology to train the cycle-consistent decomposition and composition model (\Cref{fig:data_collection}(c)).
In each iteration, we first generate a large batch of composite images. The cycle model from the previous iteration is then used to perform a full decomposition-recomposition cycle on this new data.
Examples that are evaluated as high fidelity are considered pseudo-samples. These samples are retained and incorporated into the training set for each iteration. 
This self-curation process ensures that the model continuously benefits from increasingly reliable supervision, resulting in demonstrable improvements in decomposition stability, coherence, and generalizability.
}


\begin{figure*}[!t] \centering \includegraphics[width=\textwidth]{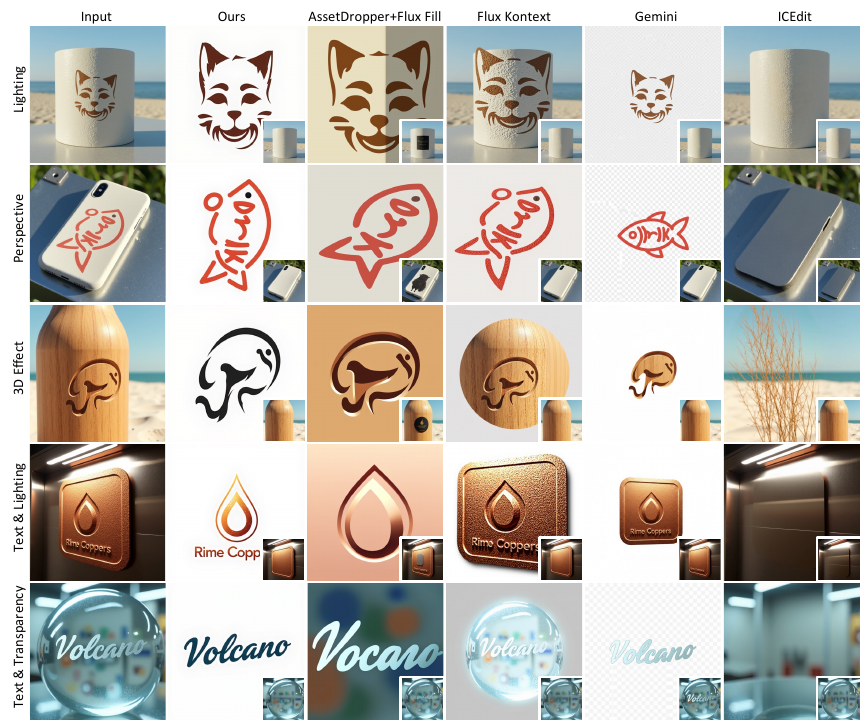} \caption{Qualitative comparison on challenging scenarios on synthetic data. The first column shows the inputs, while the following columns present results from our approach and four baselines: AssetDropper~\cite{li2025assetdropper}, Flux-Kontext~\cite{labs2025flux}, Gemini~\cite{comanici2025gemini}, and ICEdit~\cite{zhang2025context}. The decomposed object layers appear at the bottom-right of each sample. For AssetDropper~\cite{li2025assetdropper}, we use FLUX-Fill~\cite{flux2024} to inpaint the logo region as the object layer.
\gz{Note that all the synthetic images are generated from a prompt, not composited from a logo and a clean object.}
} 
\label{fig:main_comparison} \end{figure*}


\begin{table*}[!t]
\centering
\caption{Quantitative comparison using VQAScore and VLMScore, higher is better. To prevent potential evaluation bias, we utilize three VLMs as evaluators (\textit{i.e.}, Qwen, GPT-4o, and Gemini). The \textbf{best} and \underline{second} best results are highlighted.}
\resizebox{\linewidth}{!}{
\begin{tabular}{lccccccc}
\toprule
\textbf{Method} & \multicolumn{2}{c}{\textbf{VQAScore}~$\uparrow$} & \multicolumn{5}{c}{\textbf{VLMScore}~$\uparrow$ (Qwen / GPT4o / Gemini)} \\
\cmidrule(lr){2-3} \cmidrule(lr){4-8}
 & Logo & Object & Logo Isolation & Logo Consistency & Object Isolation & Object Consistency & Average \\
\midrule
AssetDropper & \underline{$0.42$} & — & \underline{$4.49$} / {\boldmath $2.87$} / {\boldmath $4.84$} & $3.84$ / $4.00$ / \underline{$3.90$} & — & — & —\\
ICEdit & $0.31$ & $0.31$ & $1.09$ / $2.70$ / $1.02$ & $1.56$ / $2.58$ / $1.00$ & $2.35$ / $3.93$ / $3.88$ & $3.80$ / $3.42$ / $3.32$ & $2.55$ \\
Kontext & $0.40$ & \underline{$0.32$} & $3.27$ / $2.73$ / $3.28$ & \underline{$4.17$} / $4.04$ / $2.80$ & $2.80$ / $3.75$ / \underline{$4.64$} & \underline{$4.75$} / $4.26$ / {\boldmath $4.94$} & $3.79$ \\
Gemini & $0.42$ & {\boldmath $0.32$} & $4.39$ / $2.63$ / $4.72$ & $4.05$ / \underline{$4.36$} / $3.72$ & \underline{$3.09$} / {\boldmath $4.39$} / {\boldmath $4.72$} & {\boldmath $4.81$} / $4.63$ / \underline{$4.93$} & \underline{$4.20$} \\
\midrule
Ours & {\boldmath $0.43$} & $0.31$ & {\boldmath $4.50$} / \underline{$2.77$} / \underline{$4.76$} & {\boldmath $4.22$} / {\boldmath $4.43$} / {\boldmath $3.94$} & {\boldmath $3.35$} / \underline{$4.33$} / $4.37$ & $4.70$ / $4.62$ / $4.67$ & {\boldmath $4.22$} \\
\bottomrule
\end{tabular}
}
\label{tab:main_comparison}
\end{table*}

\begin{figure*}[!t] 
\centering \includegraphics[width=\linewidth]{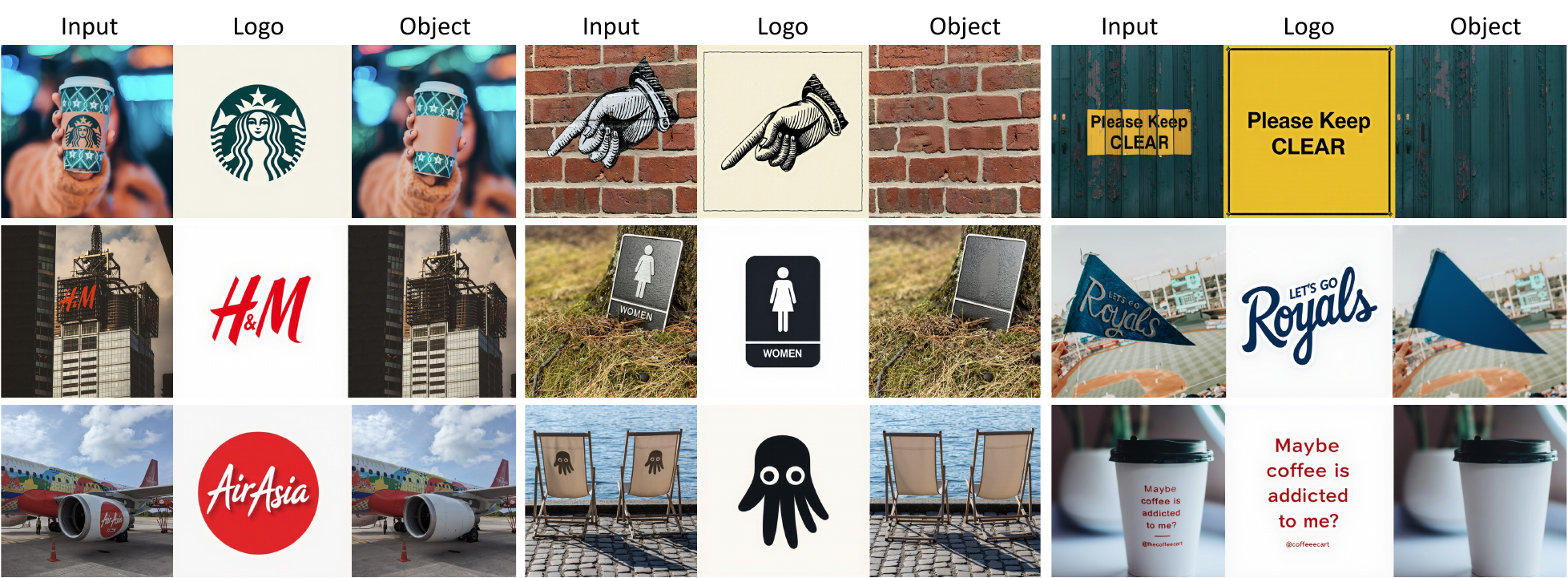} \caption{Decomposition results on real photographs. Our approach successfully decomposes both well-known brand logos and uncommon text signs under diverse lighting and perspective conditions, demonstrating strong generalization in real-world environments.} 
\label{fig:real_images} \end{figure*}

\begin{figure}[!t] 
\centering \includegraphics[width=\linewidth]{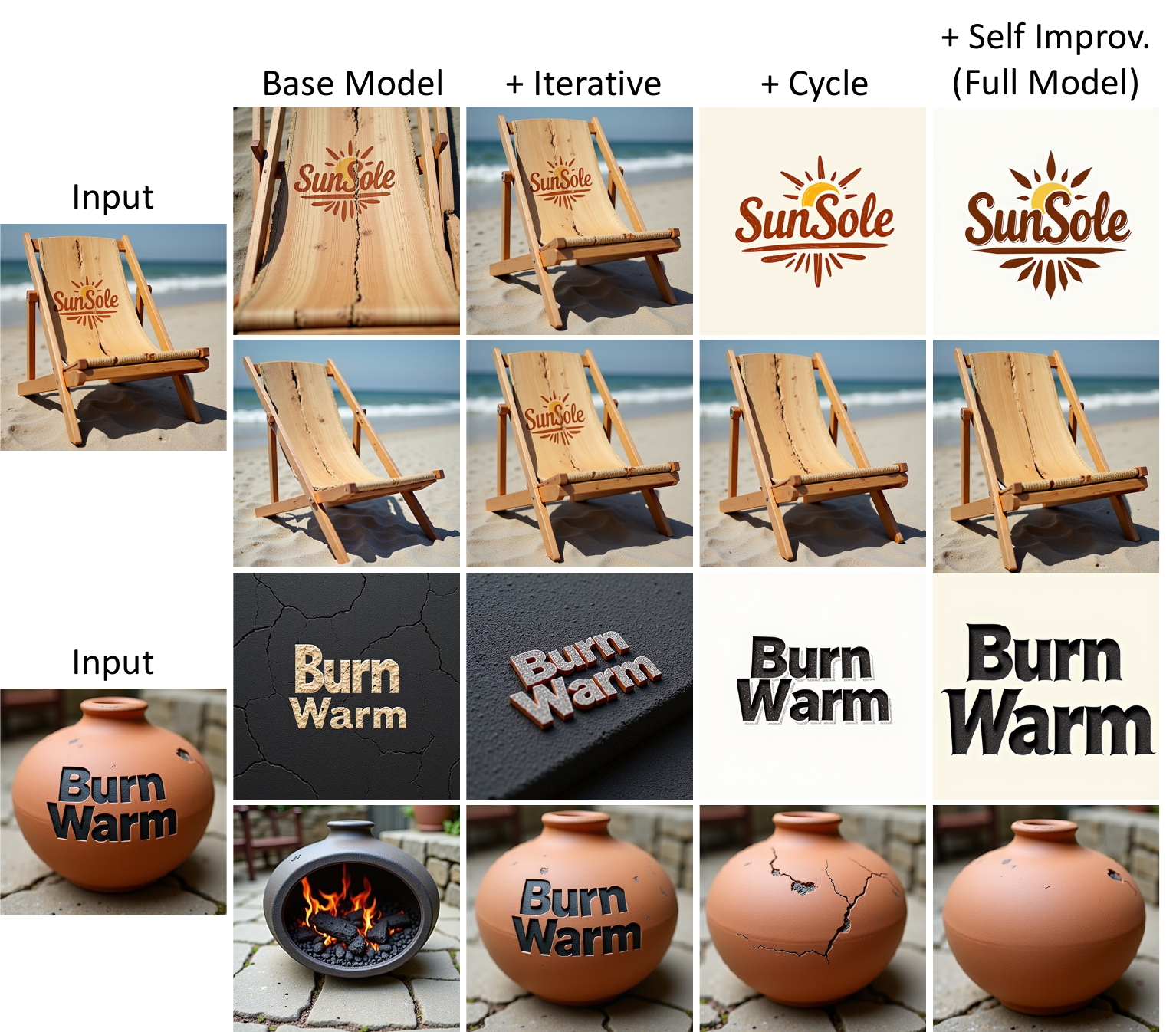} \caption{Ablation study. Each row visualizes the decomposition results with progressively adding the proposed components. 
The decomposed logos and objects become clearer and more consistent from the base model to the full model.
}
\label{fig:ablation} \end{figure}

\begin{figure}[!t] 
\centering \includegraphics[width=\linewidth]{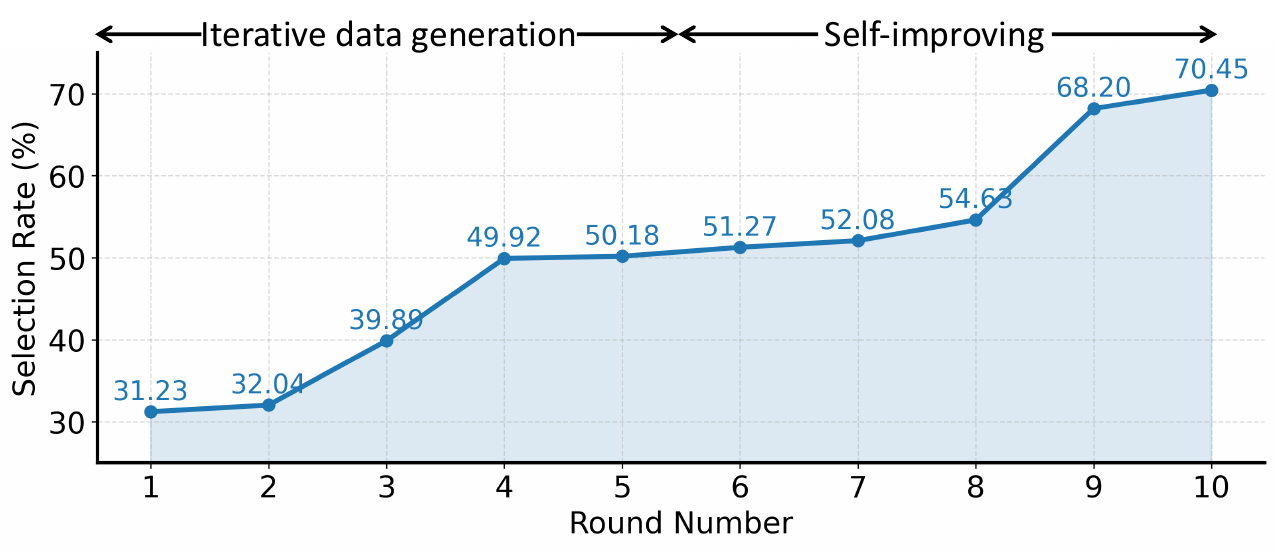} \caption{Selection rate of high-quality samples generated by our approach across different data collection rounds. 
The selection rate increases steadily as the round number grow from $1$ to $10$.
}
\label{fig:ablation_selection_rate} \end{figure}

\section{Experiments}
\label{sec:experiments}

\zs{
\subsection{Comparison Methods}

To benchmark our approach, we compare our method against baselines from two categories: asset-focused and instruction-based editing.
For asset-focused editing, we compare against AssetDropper~\cite{li2025assetdropper}, 
in which GroundingDINO~\cite{liu2024grounding} is required to locate the logo first.
For instruction-based editing, we evaluate IC-Edit~\cite{zhang2025context}, Flux-Kontext~\cite{labs2025flux}, and Gemini~\cite{comanici2025gemini}. We use the instruction ``\textit{Extract the logo, remove the object and background}'' to get the logo layer and ``\textit{Remove the logo}'' to obtain the object layer.


}



\subsection{Quantitative Results}

\zs{
Quantitative evaluation is conducted on 1.5K synthetic test samples using two metrics: 
(a) VQAScore~\cite{lin2024evaluating}, which measures text-image alignment. VQAScore is computed independently on the decomposed logo and object; 
(b) VLMScore, in which we use different VLMs to assess decomposition results on a 1-5 scale across four aspects: Logo Isolation, Logo Consistency, Object Isolation, and Object Consistency. 
Please check out the supplementary material for more detail on the evaluation.

Results in \Cref{tab:main_comparison} demonstrate that we achieve the highest logo VQAScore and VLMScore, 
confirming our superiority in separating logos. In contrast, while instruction-based models (Gemini~\cite{comanici2025gemini}, Flux-Kontext~\cite{labs2025flux}) excel at logo removal (high object scores), they struggle to accurately isolate the logo. AssetDropper~\cite{li2025assetdropper}, while specialized for asset extraction, degrades in complex non-linear scenarios and, crucially, fails to recover the underlying object.
} 



\subsection{Qualitative Results}

\zs{
We present a qualitative comparison with baseline methods in \Cref{fig:main_comparison}. The figure showcases five challenging scenarios: illumination variation, perspective distortion, logos on non-planar (3D) surfaces, textual elements, and transparent materials.
These examples demonstrate the methods' varying abilities to handle non-linear layer interactions.
Across all cases, our approach consistently produces cleaner isolated logos and more coherent objects. In contrast, competing methods exhibit noticeable artifacts, incomplete separations, or fail to preserve layer consistency. \Cref{fig:real_images} shows our method's generalization on in-the-wild photographs.

}


\subsection{Ablation Studies}


\zs{
We conduct ablation studies to validate our component design choices, with results in \Cref{fig:ablation}. 
We first establish a baseline by training the decomposition model only on data from the initial Round 0 IC-LoRA generator, which yields poor separations. 
Introducing Iterative Data Generation measurably improves decomposition, demonstrating the benefit of higher-quality data. 
Adding Cycle-Consistency provides a significant enhancement in logo fidelity and isolation. 
Finally, incorporating the Self-Improving Process (our full model) further refines object consistency and realism.
To validate our progressive data collection strategy, \Cref{fig:ablation_selection_rate} plots the selection rate of high-quality samples per round.
The increasing proportion confirms that our model's data generation quality iteratively improves, providing superior supervision for subsequent training.
}

\begin{figure*} [!t] 
\centering \includegraphics[width=\linewidth]{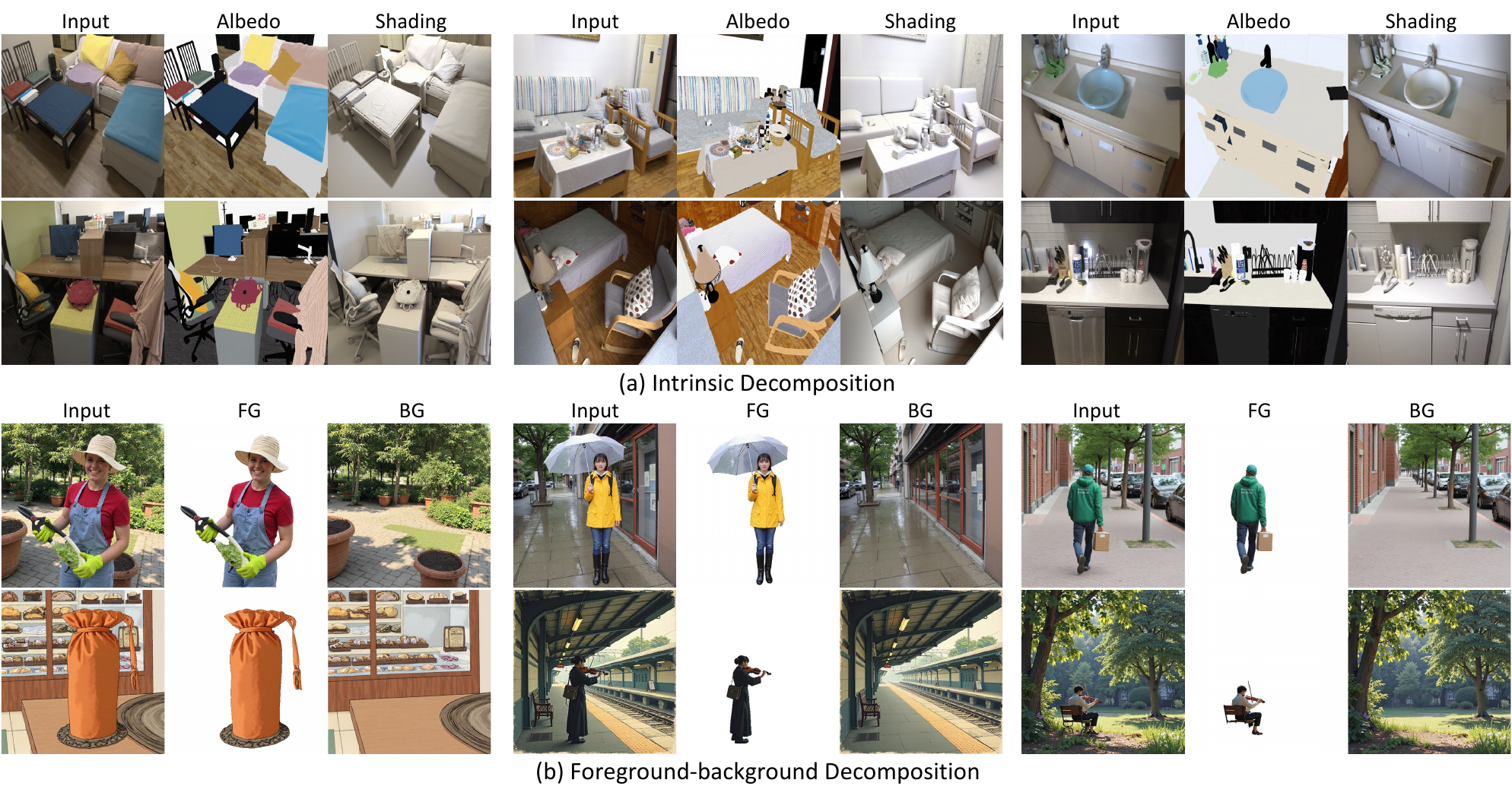} \caption{
Our decomposition results on (a) intrinsic decomposition, which decomposes an input image into albedo and shading layers, and (b) foreground-background decomposition, which separates a foreground object from its background scene. These qualitative results are supported by quantitative evaluation (e.g., \Cref{tab:intrinsic}), confirming the consistency and reliability of our decomposition framework.} 
\label{fig:results_application} \end{figure*}

\subsection{User Study}
To complement VLM-based evaluation, we conducted a user study consisting of 20 questions with 30 participants. In each question, participants ranked four anonymized decomposition results from best to worst according to two criteria: (a) consistency, measuring how accurately the logo is extracted, and (b) perceptual reasonableness, assessing whether the result appears natural without non-linear artifacts. The results are summarized below.

\begin{figure}[t]
\centering \includegraphics[width=\linewidth]{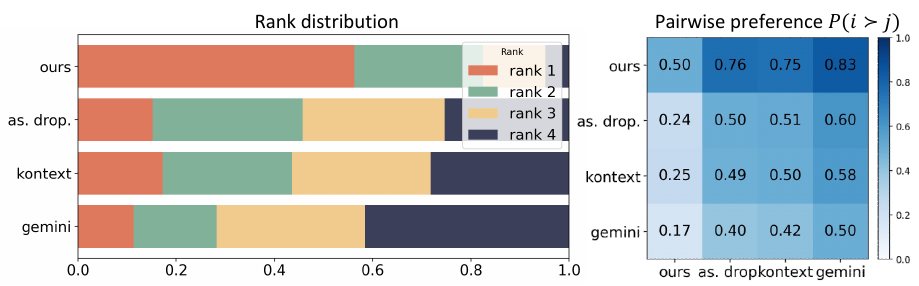}
\caption{Visualization of user study results. Left: Rank distribution; Right: Pairwise preference heat map.}
\label{fig:user_study}
\end{figure}

As illustrated in \Cref{fig:user_study}, our method is ranked top-1 in over 50\% of the cases. AssetDropper~\cite{li2025assetdropper}, as a task specific logo extraction method, outperforms Flux-Kontext~\cite{labs2025flux} and Gemini~\cite{comanici2025gemini}, which is consistent with its design focus. While Gemini~\cite{comanici2025gemini} achieves strong VLM scores, it does not show the same advantage in user study, likely due to tendencies such as preserving original size or producing pseudo-transparent outputs, which are less favored by human.

\zs{
\subsection{Generalization to Other Decomposition Tasks}

}


\paragraph{Intrinsic Decomposition.}
\zs{
We evaluate our method on intrinsic decomposition, which factors an image into reflectance (albedo) and shading.
Without task-specific priors, we train our cycle model on the Hypersim dataset~\cite{roberts2021hypersim} and evaluate on the MAW dataset~\cite{wu2023measured}. As shown in \Cref{tab:intrinsic}, our approach achieves comparable accuracy to SOTA methods specifically tailored for this task. Crucially, an ablation against a variant lacking cycle consistency shows consistent improvement, demonstrating its benefit in stabilizing this decomposition. Qualitative results in \Cref{fig:results_application}(a) further verify our framework's effectiveness.
}


\begin{table}[t]
\centering
\caption{Quantitative comparison on intrinsic decomposition. For the compared methods, we report their results presented by Careaga et al.~\cite{careaga2024colorful}. Lower values indicate better performance.
}
\resizebox{0.9\linewidth}{!}{
\begin{tabular}{lcc}
\toprule
\textbf{Method} & \textbf{Intensity ($\times$100)$\downarrow$} & \textbf{Chromaticity $\downarrow$} \\
\midrule
Careaga et al.~\cite{careaga2023intrinsic} & \underline{0.57} & 6.56 \\
Kocsis et al.~\cite{kocsis2024intrinsic} & 1.13 & 5.35 \\
Chen et al.~\cite{chen2024intrinsicanything} & 0.98 & 4.12 \\
Careaga et al.~\cite{careaga2024colorful} & \textbf{0.54} & \textbf{3.37} \\
\midrule
Ours (w/o Cycle) & 0.59 & 3.65 \\
Ours & \underline{0.57} & \underline{3.54} \\
\bottomrule
\end{tabular}
}
\label{tab:intrinsic}
\end{table}

\paragraph{Foreground-background Decomposition.}

\zs{We also test our method on foreground-background decomposition}, which aims to disentangle salient objects from their contextual environments. 
\zs{We make a training set of $\sim$5K triplets by generating composite images (FLUX.1-dev~\cite{flux2024}), creating the background layer (Flux-Kontext~\cite{labs2025flux}), and extracting the foreground (GroundingDINO~\cite{liu2024grounding} and SAMv2~\cite{samv2}). After training our cycle model on this dataset, it produces clear decompositions, as shown in \Cref{fig:results_application}(b). These results further demonstrate the generalization of our framework to other separation tasks.}


\section{Conclusions, Limitations, and Future Work}
\label{sec:conclusion}

In this paper, we revisited the problem of layered image decomposition through the lens of in-context learning. While diffusion-based in-context models have mostly been explored for composition and editing, we showed that the same paradigm can also be extended to the inverse process to decompose or unmix an image into its constituent layers. Central to our framework is a cross-cycle training scheme, where decomposition and composition models are trained jointly to validate each other, enabling the handling of non-linear and globally coupled interactions between layers through one model. This is different from previous work DecompDiffusion~\cite{su2024compositional}, which addresses image decomposition by training separate models.

While our method shows strong and consistent performance across a wide range of settings, it is not entirely bulletproof. The model still struggles with certain out-of-domain cases, particularly when the overlaid element dominates the scene, such as very large brand logos on walls or billboards. These cases often fall outside the distribution of training data and remain challenging for current models trained under limited diversity. Another limitation is the handling of multi-layer decomposition. Our current formulation is designed for separating up to \gz{two} layers and cannot easily generalize to images containing multiple overlaid elements, such as posters with several distinct logos or text layers. This constraint mainly arises from the \gz{limited-grid paradigm} used for inpainting-based visual in-context learning, a limitation shared with other VICL approaches~\cite{bar2022visual,zhang2025context}.

Beyond these limitations, the formulation reflects a broader idea that generative models can learn not only to compose but also to disassemble. Treating composition and decomposition as dual, cross-linked processes allows the model to internalize how image layers interact, rather than relying on explicit supervision or handcrafted priors.

Looking ahead, coupling decomposition and composition through mutual supervision opens several promising directions. The same principle may extend to motion, illumination, or multimodal information such as audio and 3D structure. More broadly, it encourages models that discover structure from weak or implicit supervision, moving toward a unified understanding of visual composition.

\section*{Acknowledgements}
This work was supported in parts by Guangdong Science and Technology Program (2024B0101050004), ICFCRT (W2441020), Shenzhen Science and Technology Program (KJZD20240903100022028, KQTD20210811090044003, RCJC20200714114435012), NSFC (62472288), GD Natural Science Foundation (2026A1515010423), Israel Science Foundation (3441/21, 1473/24, 2203/24), and Scientific Development Funds from Shenzhen University.

{
    \small
    \bibliographystyle{ieeenat_fullname}
    \bibliography{main}
}

\clearpage
\renewcommand{\thesection}{\Alph{section}}
\renewcommand{\thefigure}{S\arabic{figure}}
\setcounter{page}{1}
\setcounter{section}{0}
\setcounter{figure}{0}
\maketitlesupplementary


\section{Implementation Details}

\subsection{Training and Inference Details}

We implement our approach with PyTorch.
Our data curation involves a 5-round iterative generation stage ($\sim$5k samples) followed by a 5-round self-improving stage ($\sim$5k samples), yielding a final corpus of $\sim$10k samples.

In each round of iterative data collection, the model is optimized for 4K steps with the learning rate set to 1. The LoRA rank and alpha is set to 32 for the IC-LoRA adapter.
We train the cycle-consistent model for 5K to 10K steps in each round, with the step count increasing across rounds as the dataset grows. The learning rate gradually decays from 1 to 0.5. The LoRA rank and LoRA alpha are also set to 32. 

We use Qwen-VL-7B~\cite{bai2025qwen2} as a VLM-based filter to evaluate and select high-quality generated examples.
Inspired by Zhang et al.~\cite{zhang2024transparent}, the newly added high-quality samples in each round are assigned double the sampling weight.

Experiments are conducted on eight NVIDIA L40 GPU. The inference takes 35 seconds per image with 50 steps, which is within the normal inference time of Flux-Fill. Notably, our method does not require multiple runs. The logo and object are produced together in one generation.

\subsection{Cycle Consistency Loss}

In our proposed cycle consistency loss, consider track 1, the model first performs decomposition and then composition. The outputs from the first stage serve as the inputs to the second stage. Since flow-based Diffusion Models do not directly predict images but velocity fields instead, we must convert the first-stage outputs into a clean latent required for the second-stage. 
To achieve this, in our implementation, we approximate a clean latent after the first stage using the following procedure:
\begin{equation}
    \overline{x}^0 \approx x_t - t \cdot v_{\theta} \left( \left[ x_t, \mathcal{E}_{img}(X), M_d \right], t, \mathcal{E}_{txt}(T_d) \right),
\end{equation}
where $v_{\theta}$ denotes the model, $\mathcal{E}_{img}$ and $\mathcal{E}_{txt}$ are image and text encoders, $M_d$ and $T_d$ are binary mask and text prompts for the decomposition task.
Below we show that this approximation is reasonable.

During each training step, a random timestep $t \in [0,1]$ is sampled to construct a noisy latent by linearly mixing the clean latent $x_0$ and the random noise $x_1$ by:
\begin{equation}
    x_t = (1-t)\cdot x_0+t\cdot x_1.
\label{eq:xt}
\end{equation}
The model then predicts the velocity $v$ conditioned on $x_t$, together with auxiliary inputs including the masked image, the binary mask, etc. The objective is encouraging the model prediction $v$ to match the direction between $x_1$ and $x_0$ by minimizing the following term:
\begin{equation}
    \mathcal{L} = ||v - (x_1-x_0)||^2_2.
\label{eq:loss}
\end{equation}
When this objective becomes sufficiently small (i.e., closely heading zero), combining \Cref{eq:xt} and \Cref{eq:loss} yields an approximation of the clean latent:
\begin{equation}
    x_0 \approx x_t - t \cdot v.
\end{equation}
Given this approximation, we can recover an estimate of the clean latent and feed it into the second stage, which is both computationally efficient and empirically stable.
In practice, we sample two timesteps independently for track 1 and track 2 to strengthen the cycle constraints from different noise levels.
\Cref{alg:cycle} provides a Pytorch-like pseudo-code to calculate the cycle consistency loss in one step.\footnotemark
\footnotetext{In this pseudo-code, diffusion latents $x_0, x_1, x_t$ are written as $x^0, x^1, x^t$ to reserve subscripts for other indices.}

\begin{algorithm}[t]
\caption{Pytorch-like pseudo-code for the calculate of cycle-consistent loss in one training step}
\label{alg:cycle}

\SetKwComment{Comment}{\textcolor{algcomment}{\#\ }}{}
\SetKwFunction{GetPred}{get\_pred}
\SetKwFunction{CycleStep}{cycle\_step}
\SetKwFunction{EncImg}{enc\_img}
\SetKwFunction{Transformer}{transformer}
\SetKwFunction{mse}{mse}
\SetKwProg{Fn}{Function}{}{end}

\Comment*[l]{\textcolor{algcomment}{$X_{1 \times 3}$: ground truth image}}
\Comment*[l]{\textcolor{algcomment}{$M_d, M_c$: binary masks}}
\Comment*[l]{\textcolor{algcomment}{$\tau_d, \tau_c$: text embeddings}}
\;

\Fn{\GetPred{$X, M, t, \tau, x^{0}{=}\text{None}, x^{1}{=}\text{None}$}}{
    \Comment*[l]{\textcolor{algcomment}{prepare latents}}
    \If{$x^{0}$ is \text{None}}{
        $x^{0} \leftarrow \EncImg \left(X \right)$
    }
    \If{$x^{1}$ is \text{None}}{
        $x^{1} \leftarrow \mathcal{N} \left(0,I \right)$
    }
    $x^{t} \leftarrow  \left(1-t \right) \cdot x^{0} + t \cdot x^{1}$\;
    $c \leftarrow \left[\EncImg \left(M \odot X \right) , M \right] $\;
    \;
    
    \Comment*[l]{\textcolor{algcomment}{predict velocity}}
    $\text{pred} \leftarrow \Transformer \left(x^{t}, c, t, \tau \right)$\;
    $\text{tgt} \leftarrow x^{1} - x^{0}$\;
    
    $\overline{x}^{0} \leftarrow x^{t} - t \cdot \text{pred}$\;
    \;
    \Return $ \left(\text{pred}, \text{tgt}, \overline{x}^{0}, x^{1} \right)$\;
}

\;

\Fn{\CycleStep{$X_{1 \times 3}, M_d, M_c, \tau_d, \tau_c$}}{

    \Comment*[l]{\textcolor{algcomment}{sample timesteps}}
    $t_1, t_2 \leftarrow \sigma \left(\mathcal{N} \left(0,1 \right) \right)$\;
    \;

    \Comment*[l]{\textcolor{algcomment}{track 1: decompose}}
    $ \left(p_d, g_d, \overline{x}^{0}_d, x^{1}_d \right) \leftarrow \GetPred \left(X_{1 \times 3}, M_d, t_1, \tau_d \right)$\;
    \Comment*[l]{\textcolor{algcomment}{track 2: compose}}
    $ \left(p_c, g_c, \overline{x}^{0}_c, x^{1}_c \right) \leftarrow \GetPred \left(X_{1 \times 3}, M_c, t_2, \tau_c \right)$\;

    \Comment*[l]{\textcolor{algcomment}{track 1: decompose \& compose}}
    $ \left(\tilde{p}_c, \cdot, \cdot, \cdot \right) \leftarrow \GetPred \left(X_{1 \times 3}, M_c, t_2, \tau_c, \overline{x}^{0}_d, x^{1}_c \right)$\;    
    \Comment*[l]{\textcolor{algcomment}{track 2: compose \& decompose}}
    $ \left(\tilde{p}_d, \cdot, \cdot, \cdot \right) \leftarrow \GetPred \left(X_{1 \times 3}, M_d, t_1, \tau_d, \overline{x}^{0}_c, x^{1}_d \right)$\;
    \;

    \Comment*[l]{\textcolor{algcomment}{reconstruction loss}}
    $\mathcal{L}_{\text{rec}} \leftarrow \mse \left(p_d, g_d \right) + \mse \left(p_c, g_c \right)$\;
    \Comment*[l]{\textcolor{algcomment}{cycle consistency loss}}
    $\mathcal{L}_{\text{cyc}} \leftarrow \mse \left(p_d, \tilde{p}_d \right) + \mse \left(p_c, \tilde{p}_c \right)$\;
    \;

    \Return $\mathcal{L}_{\text{total}} 
            \leftarrow \mathcal{L}_{\text{rec}} + \mathcal{L}_{\text{cyc}}$\;
}

\end{algorithm}

\section{LLM/VLM Prompts and Templates}

\subsection{Seed Data Generation}
Given a isolated logo image and a clean object image (which is also synthetic), we use the following instruction to generate the seed dataset under the assistance of GPT-4o.
\begin{quote}
\small\itshape
I will give you two images, showing a logo, and a product, respectively. Your task is to generate the following image: an image of the product with logo printed. \{Logo Image\} \{Object Image\}.
\end{quote}

\subsection{Iterative Data Generation}

\paragraph{LLM-Generated Prompts.}
In the iterative data generation process, we adopt IC-LoRA~\cite{huang2024context} to collect sufficient $1\times3$ grid-like images. After one training round is finished, we use the following instruction to collect a large and diverse set of structured text prompts using Qwen3~\cite{yang2025qwen3}.

\begin{quote}
\small\itshape
You are a text-to-image prompt generation assistant. Your task is to generate diverse and high-quality text prompts for branded product decomposition images.

Below are some existing examples that have been successfully generated: \{Last five generated prompts\}.

Your goal is to create a new prompt that is different from these examples in terms of: Product type/category; Brand name and style; Logo design elements; Background setting and color.

The generated prompt should maintain the format: A three-panel image grid illustrating the decomposition of a branded product.[LEFT] [MIDDLE] [RIGHT].
\end{quote}

\paragraph{Templated Prompts.}

During the self-improving process, we no longer construct new $1\times3$ ground-truth triplets beyond those obtained in the iterative data generation phase. Instead, we focus on synthesizing increasingly diverse composite images. These composites are then fed into our cycle model, which performs decomposition and recomposition to yield supervision with higher quality and diversity.

\begin{quote}
\small\itshape
A high-quality photo of a \{material\} \{function\} featuring a prominently visible logo of type \{logo type\}. 
The logo is \{logo style\}, \{logo content\}, and rendered in a \{logo color scheme\} scheme. 
It appears \{logo texture\}, positioned at the \{logo placement\} of the product, naturally following the \{geometry\} shape and \{surface state\} texture of the surface. 
The logo conforms realistically to the material — showing subtle distortions, shading, and reflections that match the underlying \{material\} surface. 
The overall composition emphasizes the tactile interaction between the logo and the object, revealing slight curvature, depth, and perspective alignment consistent with a real printed or embedded mark. 

The object is captured with a \{viewpoint\} viewpoint and is framed in a \{composition style\} composition, under \{lighting type\} lighting from the \{lighting direction\} direction, featuring \{lighting quality\} illumination in a \{environment\} setting. 
The background is \{background\}. 
The exposure is \{exposure\} with a \{color balance\} color balance. 
This image is sourced from a \{data source\}, showcasing realistic physical integration of the logo with the product surface.
\end{quote}


\subsection{VLMScore Evaluation}

We adopt Qwen2.5-VL~\cite{bai2025qwen2} to evaluate the decomposition results automatically using the following instruction.

\begin{quote}
\small\itshape
Your task is to analyze a 1x3 grid image containing three square images arranged in one row (left, middle, right).

Answer these four questions by outputting scores ranging from 1-5 (1 for "No", 2 for "Probably No", 3 for "Uncertain", 4 for "Probably Yes", and 5 for "Yes"):

1. Does the middle image show only a logo with no product visible?\\
2. Is the logo in the middle image the same as the logo visible in the left image?\\
3. Does the right image show only a product with no logo visible?\\
4. Is the product in the right image identical to the product shown in the left image (same item, shape, design, structure)?

\end{quote}

The results for the four questions above are averaged over the entire test set and reported as quantitative metrics for logo isolation, logo consistency, object isolation, and object consistency, respectively.

\section{Additional Results}

\subsection{Decompose\&Compose}

\begin{figure}[!t] \centering \includegraphics[width=\linewidth]{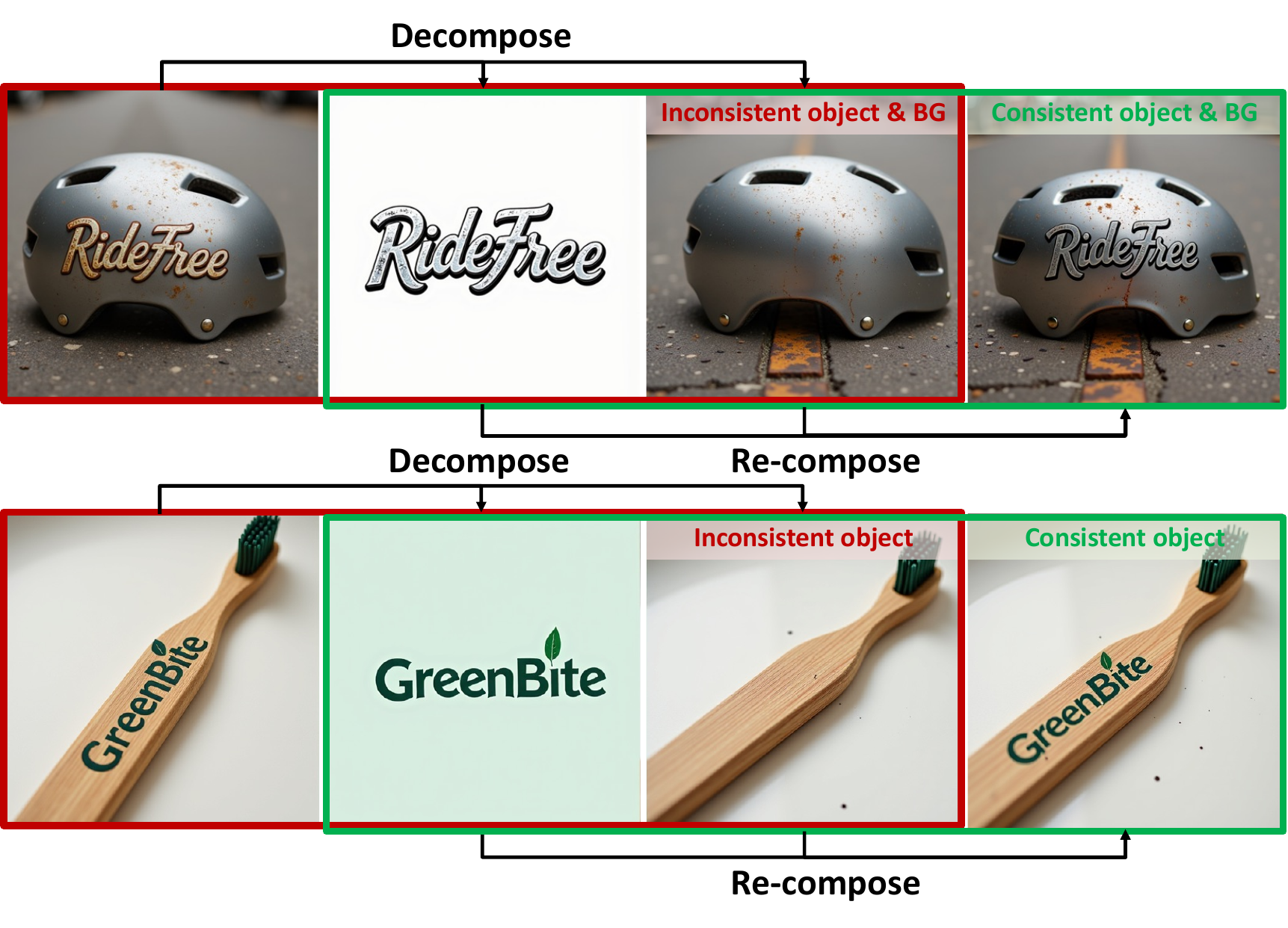} \caption{Illustration of the decompose\&compose operation. With the bidirectional generation capability of the cycle model, we are able to collect high-quality samples with improved consistency.
} 
\label{fig:supp_recomposition} \end{figure} 

\Cref{fig:supp_recomposition} illustrates the effectiveness of decompose \& compose operation in the self-improving process. Before the cycle model is introduced, the generated samples may exhibit inconsistent decomposition results (red boxes), which lead to imperfect supervision signals. After enabling the re-composition, the obtained grid provides consistency logo and objects (green boxes), allowing us to create higher-quality samples to refine the training set.

\subsection{Providing Additional Context to Baselines}
\begin{figure}[!t] \centering \includegraphics[width=\linewidth]{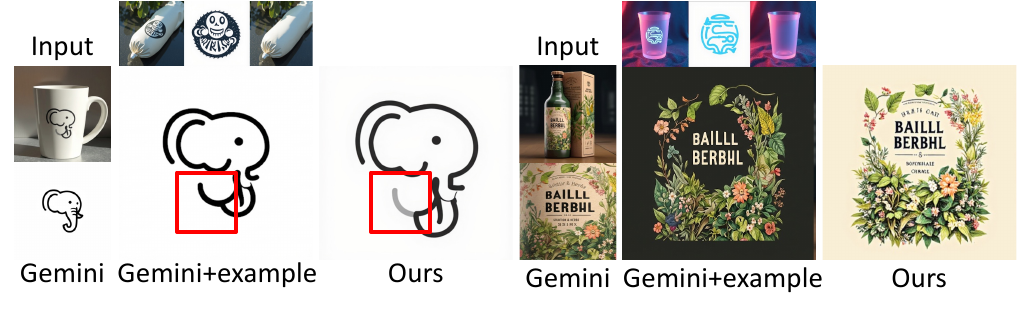} \caption{Comparison with Gemini with additional example.
} 
\label{fig:supp_gemini_with_demonstration} \end{figure} 
In \Cref{fig:supp_gemini_with_demonstration}, we evaluated Gemini both with and without an in-context demonstration, and observe that providing additional context improves its decomposition quality. Importantly, even under this strengthened setting, our method continues to produce more coherent decompositions in the presence of non-linear layer interactions.

\subsection{Comparison with Baselines on Composition}

\begin{figure}[!t] \centering \includegraphics[width=\linewidth]{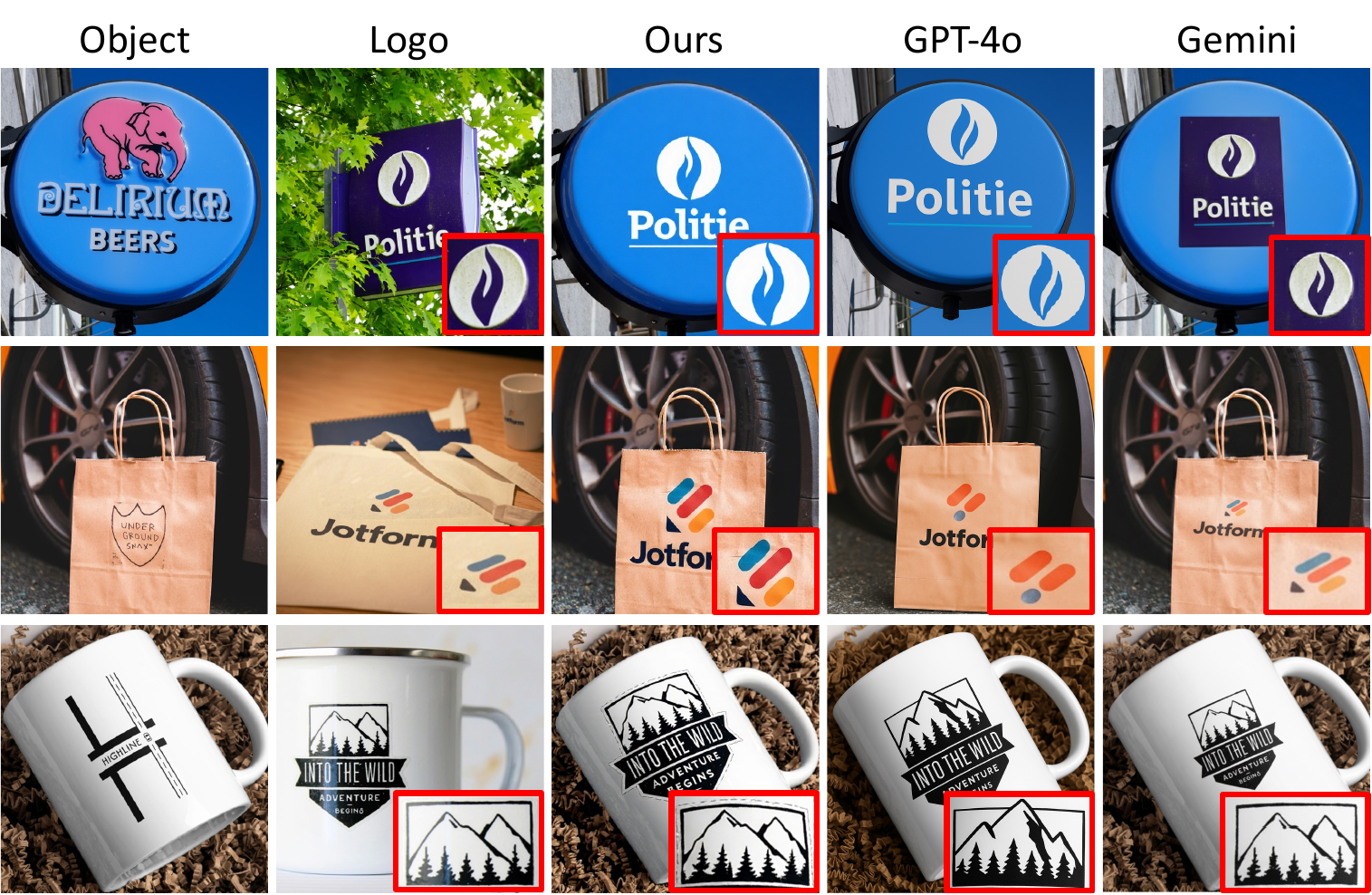} \caption{Comparison on logo-object composition. The first two rows are object and logo image, followed by outputs from Gemini, GPT-4o and our approach. 
} 
\label{fig:supp_comparison_composition} \end{figure} 

\Cref{fig:supp_comparison_composition} shows additional comparison on compositing objects and logos with Gemini and GPT-4o. Compared with Gemini, our method is less affected by object color (in the first row) and perspective distortions (in the second row). Relative to GPT-4o, we obtain more consistent logo isolation and recombination.

\subsection{Comparison with Baselines on Decomposition}

\begin{figure*}[!t] \centering \includegraphics[width=\textwidth]{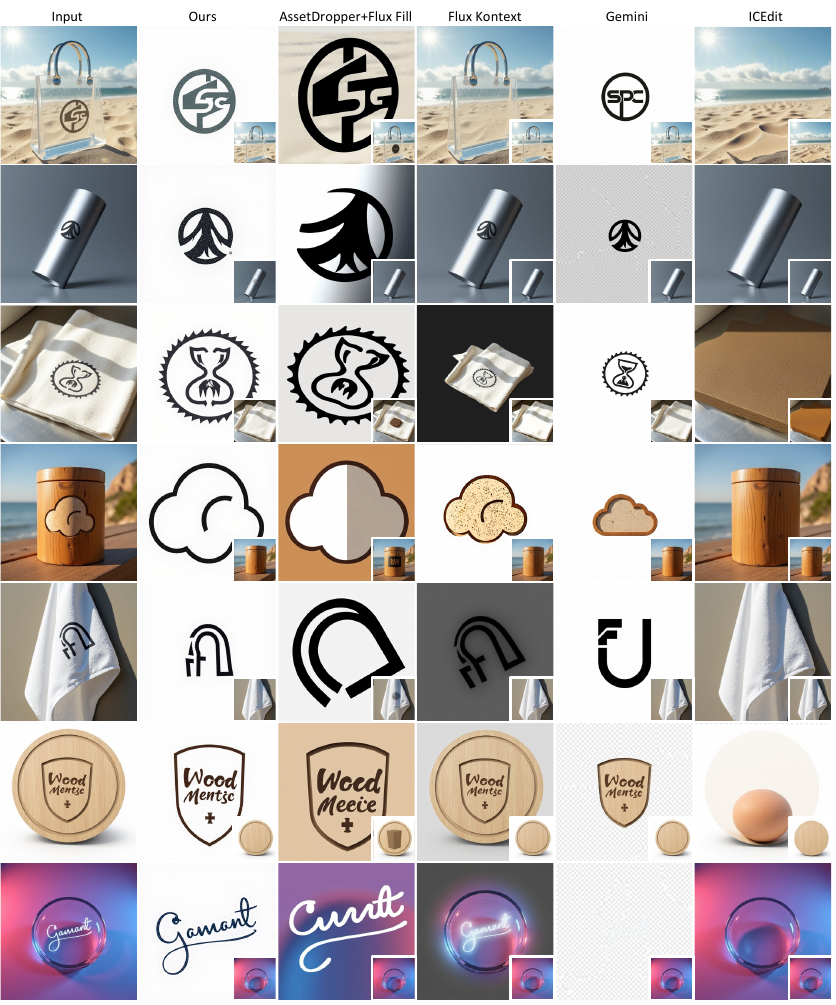} \caption{Additional qualitative comparison on challenging scenarios on synthetic data. The first column shows the inputs, while the following columns present results from our approach and baselines. The decomposed objects appear at the bottom-right of each sample. 
} 
\label{fig:supp_main_comparson} \end{figure*}

\Cref{fig:supp_main_comparson} presents additional comparison of our approach with the baseline methods. Our method produces more faithful logo and cleaner object layers under a wide range of challenging cases, including non-uniform lighting, perspective distortion, complex surfaces and transparent objects.

\subsection{Decomposition on Synthetic Images}

\begin{figure*}[!t] \centering \includegraphics[width=\textwidth]{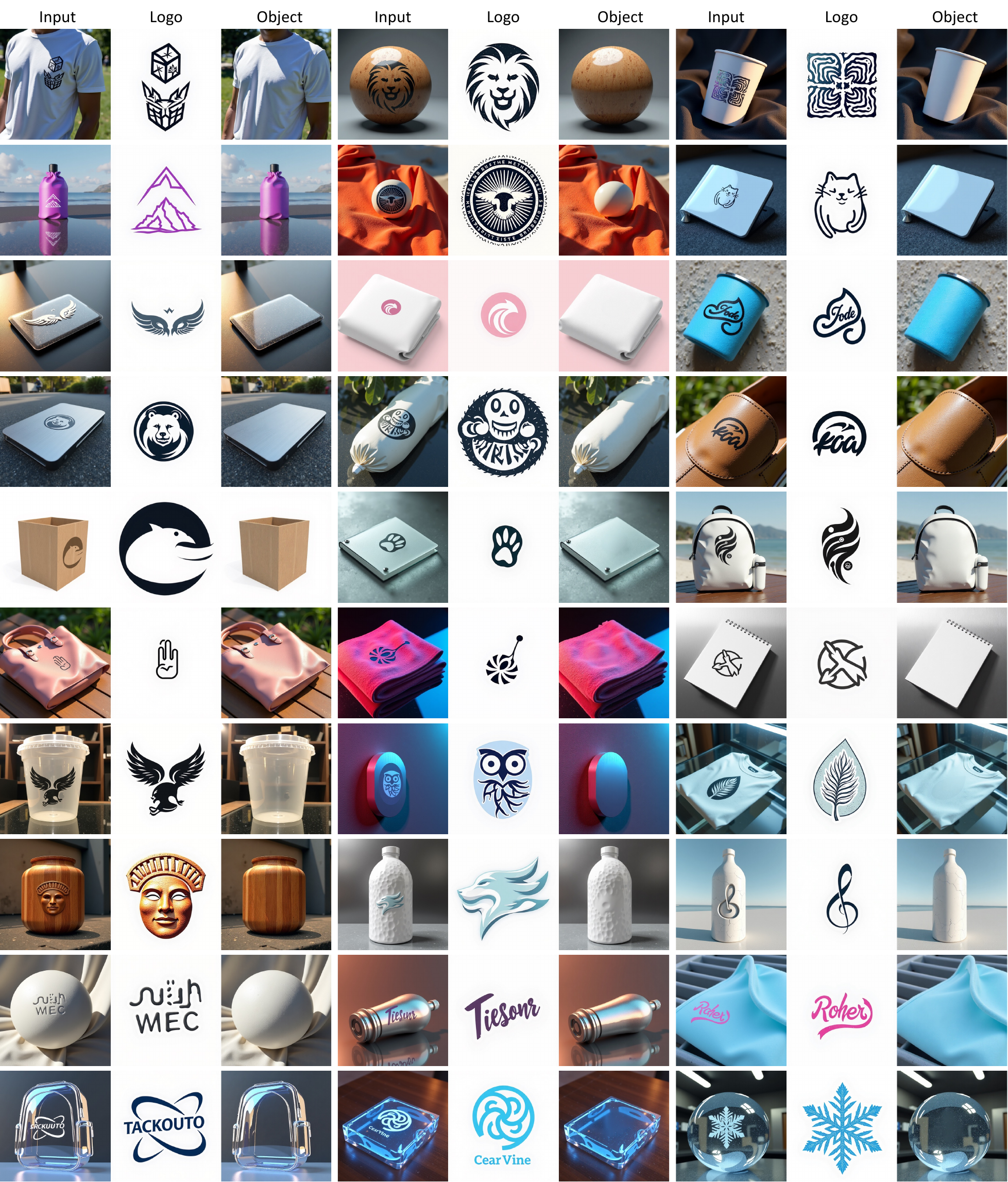} \caption{Additional decomposition results on synthetic images.
} 
\label{fig:supp_sync_images} \end{figure*}

\Cref{fig:supp_sync_images} shows additional decomposition results on our synthetic test dataset. These input images offer various challenging scenarios in lighting, perspective, material, and transparent objects. The results produce clean and consistent layers, demonstrating the effectiveness of our approach.

\subsection{Decomposition on Real-world Images}

\begin{figure*}[!t] \centering \includegraphics[width=\textwidth]{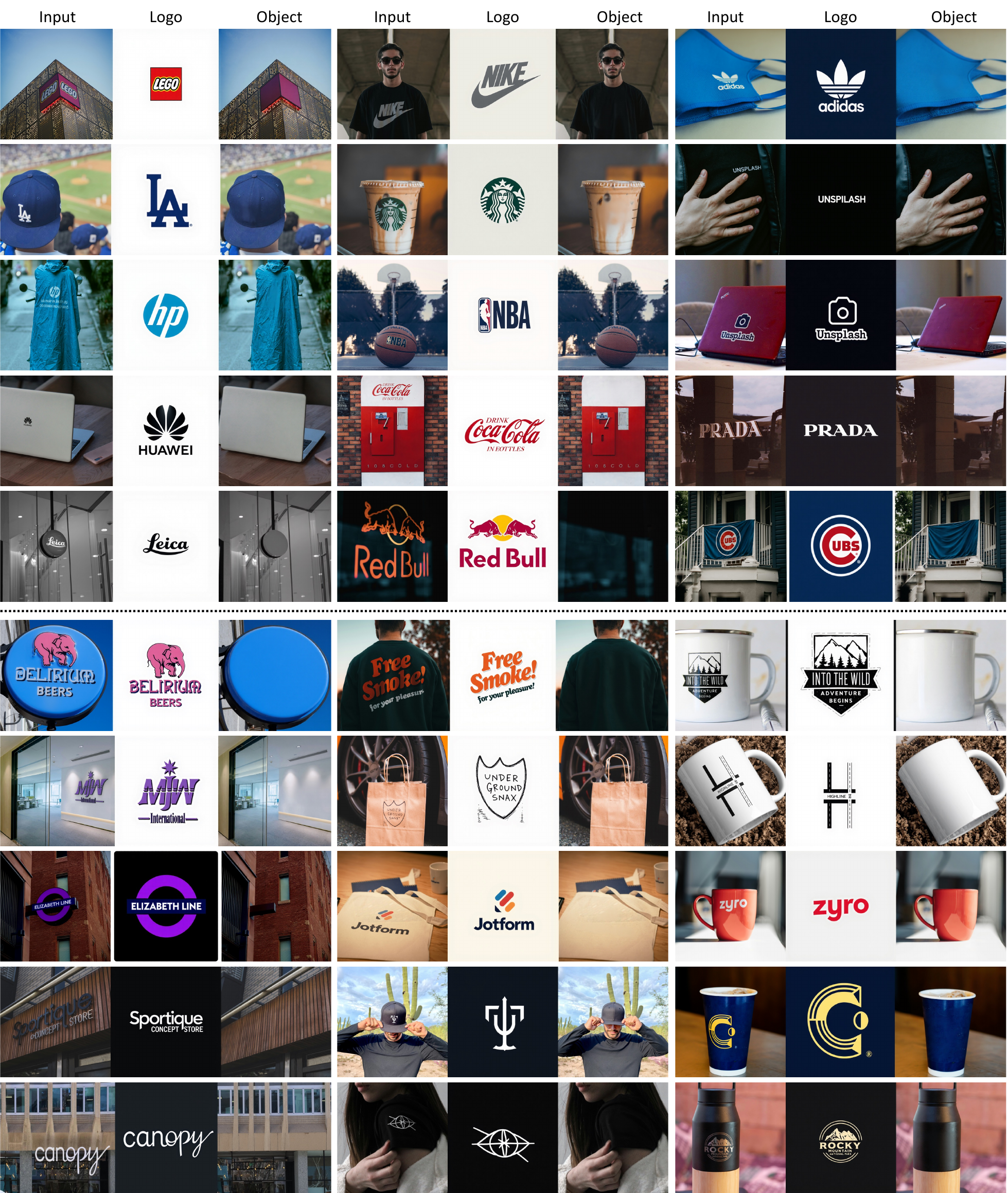} \caption{Additional decomposition results on real images. The first five rows show decomposition results for several well-known logos, while the last five rows present decomposition results for other in-the-wild logos.
} 
\label{fig:supp_real_images} \end{figure*}

\Cref{fig:supp_real_images} presents additional results on real-world photographs. In the first five rows, we show decomposition results on well-known brand logos. The next five rows show decomposition results on less common logos. 

\subsection{Intrinsic Decomposition}

\begin{figure*}[!t] \centering \includegraphics[width=\textwidth]{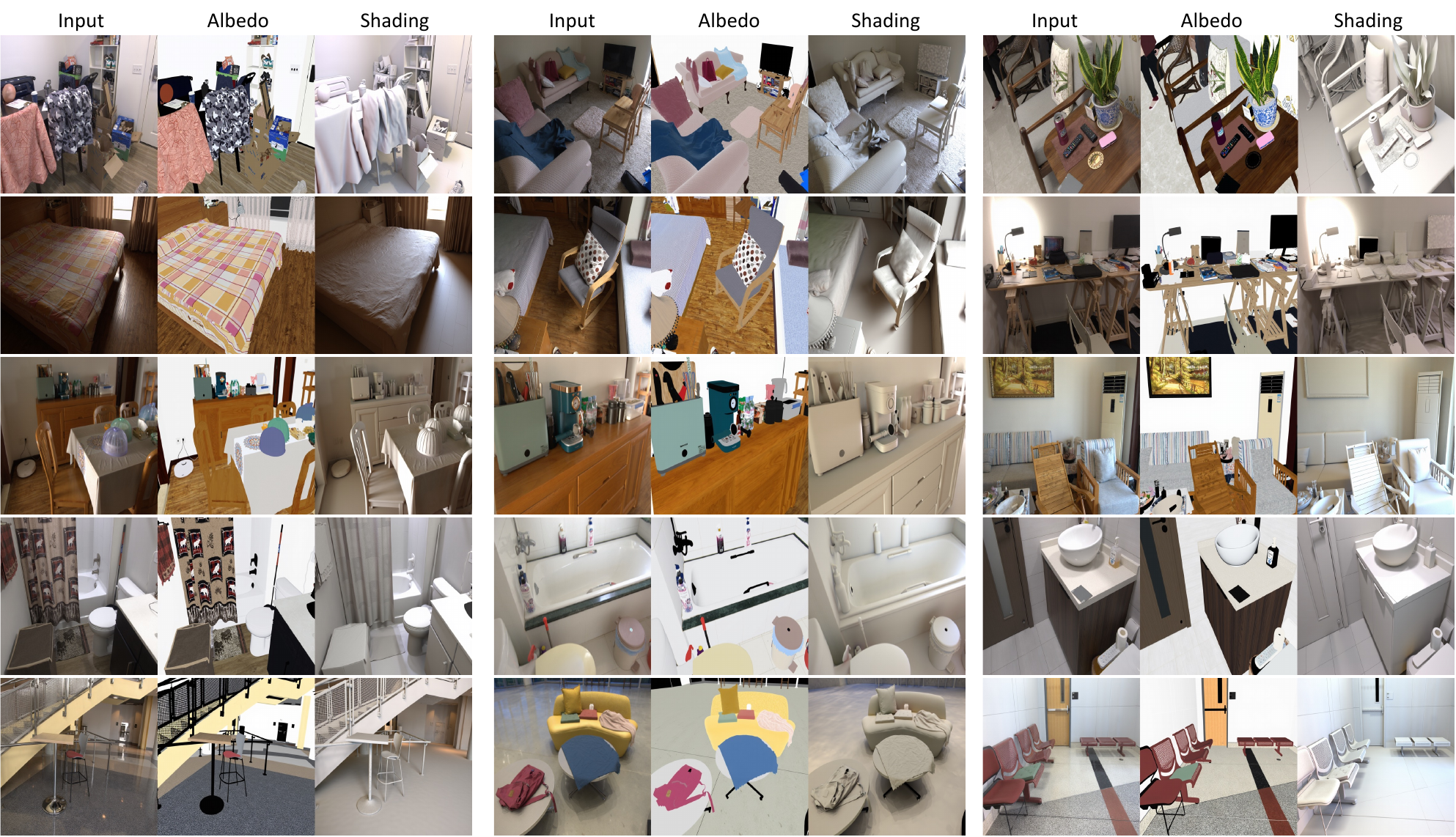} \caption{Additional results on intrinsic decomposition.
} 
\label{fig:supp_results_application_intrinsic} \end{figure*}

\Cref{fig:supp_results_application_intrinsic} shows more results on intrinsic decomposition, where our approach produces reasonable layers of albedo and shading. These results further illustrate that the same in-context generation mechanism can generalize to physics-related decompositions without hand-designed priors.

\subsection{Foreground Background Decomposition}

\begin{figure*}[!t] \centering \includegraphics[width=\textwidth]{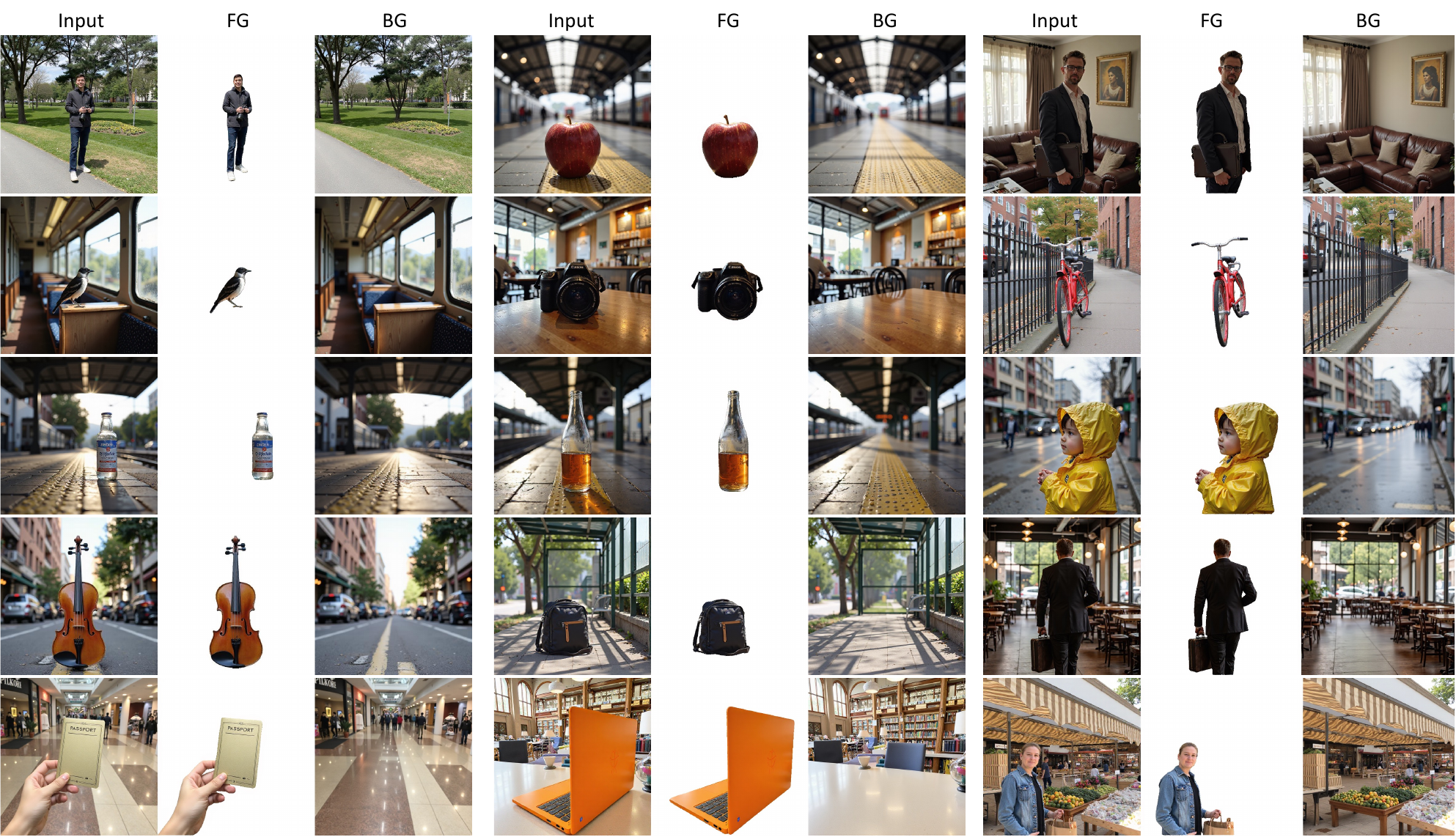} \caption{Additional results on foreground-background decomposition.
} 
\label{fig:supp_results_application_fgbg} \end{figure*}

\Cref{fig:supp_results_application_fgbg} presents additional results on foreground-background decomposition. Our approach successfully isolates the salient objects while producing coherent background layers with the effects of objects (e.g., shadows) removed. We believe these results highlight the potential of large foundation models to handle graphic tasks that require contextual consistency with internal correspondence.

\end{document}